\documentclass[11pt]{article}
\usepackage{cite}
\usepackage[margin=1in,left=1in, includefoot]{geometry} 
\usepackage{graphicx} 
\usepackage{fancyhdr} 
\title{Introduction to intelligent computing unit 1}
\author{Isa Inuwa-Dutse\\ \texttt{dutsei@edgehill.ac.uk}}
\usepackage{graphicx}
\usepackage{amsmath}
\begin{document}
\maketitle


\pagenumbering{roman}
\addcontentsline{toc}{section}{Abstract}

\begin{abstract}
This brief note highlights some basic concepts required toward understanding the evolution of machine learning and deep learning models. The note starts with an overview of artificial intelligence and its relationship to biological neuron that ultimately led to the evolution of today’s intelligent models.
\end{abstract}

\cleardoublepage 


\tableofcontents 
\thispagestyle{empty} 
\pagebreak
\pagenumbering{arabic}
\cleardoublepage

\section{Introduction} 
The advent of digital computers has led to the automation of many tasks perform by human beings. Until recently, some automated tasks were solved based on direct mapping of input to output and the computer is programme to continuously follow the specified instructions. This form of problem solving may be viewed as lacking intelligence. The need for intelligent programs to tackle real life problems was the major challenge to scientists in the 1950’s. During this period scientists came up with the interdisciplinary field which is today known as the artificial intelligence ~\cite{icp23}. Main goal of AI is to automate human tasks that require intelligence such as pattern recognition, machine translation, computer vision etc. Human beings are naturally endowed with the ability to derive knowledge from their environment through careful observation to learn distinguishing features or unique patterns in objects. The ability to infer useful information from data using appropriate tools\footnote {These tools could be from statistics, probability, machine learning etc and ~\cite{icp3}} is known as pattern recognition. Pattern recognition entails the process of exploring data in order to discover associations and cause-effect relationships ~\cite{icp8}. Application of pattern recognition is found in numerous disciplines such as history, biology, geology, computer science, and electromagnetic ~\cite{icp19}. from computing perspective, pattern recognition tasks concern with identifying and classifying data features into distinct classes ~\cite{icp24}. 
 
\subsection{Artificial Intelligence} 
The field of Artificial Intelligence (AI) is a diverse field that interests many researchers from different fields. \footnote {The approaches to AI are being considered based on the three major contributing fields: Computational Psychology, Computational Philosophy and Advanced Computer Science ~\cite{icp23}}. The interdisciplinary nature of AI has led to scientists viewing the field differently but with seemingly common goal of developing an intelligent system. For instance, mimicking intelligent tasks can be achieved through what is known as connectionism that is based on modeling of biological information processing unit found in the animal neural network ~\cite{icp15} while in the domain of psychology it refers to the model of human cognitive function ~\cite{icp16}. Models based on the connectionism consist of four main parts as in biological processing unit (the neuron): processing units, activations functions, connections and connection weights. 
 
\subsection{Information processing} 
The first attempt by the AI community was to understand how information is being processed in the biological neural network and to mimic the process. The nerve cell (neuron) is the basic unit of biological neural network that is responsible for information processing ~\cite{icp4} and inspired scientists to propose and develop a replica known as Artificial Neural Network (ANN) to intelligently handle real life tasks. ANN is simply a collection of simple interconnected processing elements whose functionality is based on imitating the animal neural network ~\cite{icp16}. The resemblance of ANN to biological neural network makes it an ideal tool to use in mimicking tasks performed by intelligent agent (human being) ~\cite{icp2}. One of the important features of ANN is the ability to learn in a manner closely related to the biological neural network learning process.  
 
\subsection{Suitability of ANN in solving real life problems} 
Real life problems are so complicated that conventional programming languages cannot be used to tackle it because most real life problems are not linear\footnote{These are problems that lack specific function to establish relationship between its inputs and outputs. The input and output cannot be represented by a linear combination of input to give the output.} but nonlinear and multi-dimensional. Some of the features that make the ANN\footnote{Others include: fault tolerant; capable of performing real time operation in parallel and self organising ~\cite{icp2}} suitable in tackling real life problems include: 
\begin{itemize} 
\item \textit{ANN-based models} are capable of solving both linear and nonlinear problem: While linear problems can be formulated using specific function (to map input-to-output relationship), non-linear problems lack specific function that will establish relationship between inputs and outputs. In other words, the input and output cannot be represented by a linear combination of input to give the output; output depends not only on the input but also on other dynamic parameters. This is the typical nature of real life problems and conventional programming languages cannot be used in this case; 
\item \textit{Input-output mapping capability:} ANN models also supports direct mapping of input-output through a training technique known as the supervised learning. For the patterns in data to be learned by ANN model to the level of classifying unknown data (also known as generalisation), the network has to undergo the process of training. Models based on ANN are quite powerful in this respect; 
\item \textit{Self-organising:} When an ANN model receives data, it is capable of organising the data in its own way and uses it for the learning purpose; and 
\item \textit{Adaptability and Resilient:} ANN models are capable of adapting to changes in the environment. In terms of being resilient, ANN-based model is being noted to be resilient to noise and hardware failure ~\cite{icp16}; these two properties are mostly found in nonlinear systems in which change in the output is not always proportional to the input. This means that a change in the input is not noticeable in the output as in the case of linear systems that easily responds to changes. With linear systems, slight change in the input causes corresponding change in the output. 
\item \textit{Generalisation:} main purpose of training ANN model (a universal model with respect to the data type trained on) is to build a model that is capable of generalising important features of the training data and be able to use that knowledge in classifying future datasets. Any model that is capable of classifying unseen data [data not encountered during training regime] correctly is said to exhibit generalisation. The ability of ANN model to learn about the different features from training data and be able to utilise the features to understand previously unseen data is known as generalisation. The ability to generalise well on real life problems is challenging due to the presence of many free parameters\footnote{This refers to the parameters that have no specific value and it is independent of the network.} to be learned by the network model. These free parameters make it difficult for an ANN model to achieve good generalisation ~\cite{icp5}.  In order to achieve good generalisation, large amount of training data are required to train ANN model ~\cite{icp8}; other factors that ensure good generalisation are the introduction of constraints to the network ~\cite{icp5} and the use of training data that is less noisy\footnote{Noisy data are the data that got distorted during data collection and this distortion renders the data less reliable for use in training ANN model ~\cite{icp22}} ~\cite{icp10}. 
\end{itemize} 
\subsection{Intelligent computing unit} 
The biological neural network is the source of inspiration for scientists interested in developing an artificial replica of the biological neuron. The nerve cell (neuron) is a simple, yet powerful processing unit consisting of three basic components (figure 1) identified as soma (cell body), tubular axon (tube like component with extended branches) and dendrites (hair-like component surrounding the neuron). These components and their functions are replicated in the artificial neural network models. 
\begin{center} 
\includegraphics[scale=0.5]{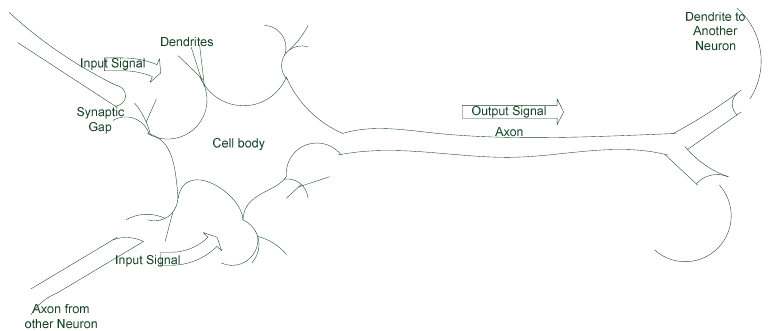}\\ 
Figure 1: A simple structure of biological nerve cell. The cell body of the nerve cell housed the nucleus that stores hereditary traits. Incoming signals are received from other neurons via the dendrites and the axon transmits generated signals to other branches. The Synapse at the terminal is the functional unit of the neuron. 
\end{center} 
Basic functional components in the nerve cell: 
\begin{itemize} 
\item \textit{Soma (cell body):} all incoming signals from other neurons are summed in the cell body. The summed signals are transmitted (i.e. the cell firing) along the tube-like axon to other cells. Signals transmission is only possible when the summed signals reach certain level otherwise no transmission. This form of transmission was replicated in ANN model response to be classified as binary, e.g. 1 when neuron transmits or 0 when no transmission; 
\item \textit{Dendrites:} the signals (electrical impulses) coming from other neurons are received by the dendrites (analogous to input layer in ANN model) and the synaptic gap found in the dendrites transmits the signal by a means of chemical process. This chemical process is capable of modifying incoming signals. The modification of incoming signals is achieved by scaling the frequency of the receiving signals ~\cite{icp17}; and 
\item \textit{Axon:} the axon is the output unit of the neuron that transmits generated signals to other branches and it form synaptic connection with other neurons. 
\end{itemize} 
 
The cell body (membrane) in the nerve cell receive multitude of incoming signals which are summed and the cell decide the output based on the summed signals. Some of the received signals exhibit certain features that affect cell’s response. \textit{Inhibitory signal} prevents the cell from responding or firing to the input signals and \textit{exhibitory signal} facilitates the generation of output signal. These two forms of signals together with the \textit{interunit connection strengths} (known as a the weights) determine whether the nerve cell will generate output signal or otherwise. With this brief overview of biological nerve cell and associated functions of components, next section focus on how artificial replica of the nerve cell is achieved. 
 
\section{Artificial Neural Network} 
Artificial neural networks (ANN) are the equivalent of the biological neural networks (in some respect) such as the collection of interconnected processing elements. One of the important features of ANN is the ability to learn from experience or training data. The basic model of the ANN that shows the potential of learning was first proposed in 1943 by McCulloch and Pitts. The Combinations of these simple processing units of ANN are powerful in solving complex problems and the parallel structure of neurons in the network contribute to computational power ~\cite{icp4}. McCulloch and Pitts model is based on threshold logic unit (TLU) that functions by comparing summed weighted-input with a threshold value and decide the outcome. If the activation value exceeds the threshold, an output of 1 is realised otherwise 0. Various forms of activation functions (see section 2.4) are used such as the signum function that uses 1 and -1 (bipolar) instead of 1 and 0 (binary). 
 
\subsection{Learning Processes in ANN} 
Basically, learning process involves adjustment to stimulant receive from the environment and responding accordingly. Learning process in ANN (figure 2) involves updating the network architecture and the network connection weights ~\cite{icp4}. 
\begin{center} 
\includegraphics{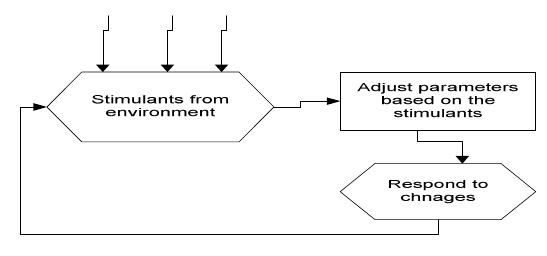} 
 
Figure 2: An abstract view of a learning processes. The three arrows represent stimulant received from the environment; these stimulants cause a change in the network parameters (weight vectors in this respect) and the model respond accordingly. The responses from the model are fed back as stimulants and the process repeats until the model understands or learns the behaviour of the environment. 
\end{center} 
 
Common learning rules used in ANN models include ~\cite{icp4}: 
\begin{itemize} 
\item \textit{Error-correction rule} this form of learning is based on error generated during the process of learning and the subsequent adjustments of network free parameters\footnote{These are parameters that keep on changing during learning process by ANN model. As dynamic parameters, they keep on changing until the appropriate response is found.}. Learning occurs only when there is error in the network (absence or very minimal error implies the model learns well about the data). Error function given by: 
\begin{equation} 
E(d,y)=\frac{1}{2}g(d-y)^2 
\end{equation} 
where E is summed over all the data samples and error, $E$, is generated whenever $y \neq d$; ($y$ is the model output and $d$ is the desired output from the model). The ultimate goal is to minimize the error function through weights modifications. 
\item \textit{Hebbian learning rule} is based on the work of Hebb 1949 ~\cite{icp14} that establishes transmission relationship between two transmitting nerve cells. Nerve cells establish strong connection if they consistently transmit signal to one another (firing one another) likewise the synaptic connection is weakened if the cells do not consistently transmits to one another. 
\item \textit{Competitive learning rule} also known as the winner-take-all, is based on competition amongst output units. The output units are simultaneously activated where the output unit with the best response is noted and attention is shifted toward it. 
\item \textit{Boltzmann Learning Rule} consists of two state of ‘on’ and ‘off’ in which weights in the network are symmetrical. 
\end{itemize} 
\subsection{Training ANN to learn} 
Learning process in ANN heavily relies on data; large amount of dataset (training data) are used to train the network to learn how to perform a specific task ~\cite{icp1}. As a mathematical model, the ANN is capable of modeling complex data to discover its relationship. The goal of training the network is to learn optimum values the network weights vectors that store the computational capability of the ANN ~\cite{icp16}. Popular training approaches can be categorised into supervised learning; unsupervised learning and reinforcement learning ~\cite{icp9}. 
\subsubsection{Supervised Learning} 
The Supervised learning method utilises network input data and its corresponding output data (also known as the desired output/target) to train the network or model on. The network input data are processed and result compare with the supplied output data [desired]; difference between the desired output, d and actual output, y is taken and adjustments are made to the weight vectors based on the difference [this difference $(d-y)$ is considered as the network error]. The error is propagated back through the network and the process repeats until the desired result achieved [through error minimisation]. Perceptron and Multilayer Perceptron learning algorithms are based on this training method. Classification tasks are based on the supervised learning techniques 
\subsubsection{Unsupervised learning} On the contrary, the unsupervised learning method requires no input and output pairs. This method requires only input vectors and the learning regime to decides on the feature to use when self- organising the weights vectors. Clustering tasks are based on this learning technique 
\subsubsection{Reinforcement learning} Unlike the supervised and unsupervised learning methods, the reinforcement learning method is not directed on what to do; it generates its data based on interactions with the environment since no data is supplied ~\cite{icp17}. This method identifies the characteristics of the problem to solve and act based on it. Essentially, this technique is based on mapping situations to actions. 
\subsection{ANN Architecture} 
ANN architecture is simply the way connection patterns (topologies) are structured in the network with neurons structured in directed graph like fashion. This architecture is used to place an input pattern into one of the different classes based on the output pattern ~\cite{icp16}. Structure of ANN is one of the determining factors\footnote{Other factors on which performance depends include: learning algorithm (and corresponding parameters) and number of iteration during training of model.} of how best ANN model perform ~\cite{icp18}. For complex problems to be solved by ANN, large assembly of interconnected artificial neurons are needed. Two common types of architectures commonly used in ANN model and architectures are differentiated based on topologies. 
\subsubsection{Feedforward architecture} 
The architecture of feedforward network (kind of acyclic graph) consists of layers with unidirectional connections through which network signal flows from the beginning to the end where it is transformed into the network output. The Perceptron is a simple form of ANN model that is based on this architecture. When the layer in the Perceptron is increased, a new ANN model known as Multi-Layer Perceptron (MLP) is produced. Since the MLP introduces additional layer to the perceptron architecture. Then the MLP is always made up of at least one hidden layer (figure 3). 
\begin{center} 
\includegraphics[scale=0.5]{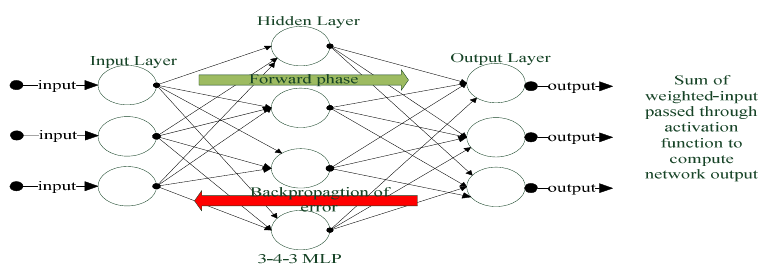}\\ 
Figure 3: A simple feedforward multilayer Perceptron network with 3-nodes input layer; 4-nodes hidden layer and an output layer with 3-nodes. Weighted-input signal flow from the input layer to the output layer and the error is propagated back from the output layer to input layer where weight modifications are made based on the back-propagated error. Approaches in deep learning rely numerous hidden layers to accomplish complex tasks previously known to be difficult to handle such as computer vision. 
\end{center} 
\subsubsection{Feedback architecture} 
Conversely, feedback or recurrent network architecture (also similar to cyclic graph) comprise of loops that permit feedback from the network to itself. The flow of signal in this form of architecture is bidirectional (signals flow in both forward and backward direction). 
\subsection{Activation Functions} 
The goal of ANN learning algorithm is to minimise generated errors, \textit{e} that are based on the network response, y. The activation function is responsible for the network response; it specifies the output of a given input data because the response from the summation unit alone will not yield useful result by plotting response from the summation unit (sum of weighted-input) results in a straight line that is insufficient to describe the data. Introduction of activation function appropriately decide the network/model response. Common activation functions in ANN models are the step function and the sigmoid function. Others include \textit{rectified linear unit (relu), softmax}. Details to be provided in part 2. 
\subsubsection{Step Function} 
ANN models such as the TLU and Perceptron use the step function as the activation function to generate a nonlinear response (figure 4) that limits the output to binary response (1/0) or bipolar response (1/-1). 
\begin{center} 
\includegraphics{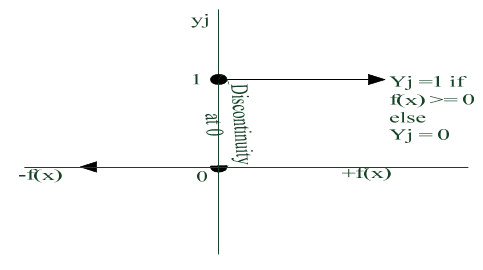}\\ 
Figure 4: A simple step function in which $f(x)$ is unbounded while the response $y_j$ is restricted to take on only binary value of either 1 or 0. The function also illustrates the discontinuity of the function at $f(x) = 0$. 
\end{center} 
The discontinuity of the step function at $f(x) = 0$ makes it unsuitable when dealing with nonlinear problems such as the simple logic XOR. Another form of step function is the signum or sign function in that takes on the value of -1 or +1 (bipolar) instead of 0 or 1 (binary). 
\subsubsection{Sigmoid Function} 
The complexity of real life problems goes beyond the capability of simple step function. The difference between these activation functions is in using the summed weighted-input ($\Sigma w_i x_i$) with a different activation function. Unlike the step function that suffers from discontinuity at $f(x) = 0$, the sigmoid function (figure 5) is continuously differentiable (bounded by 1 and 0). This differential property makes it suitable for use in ANN models to solve complex and nonlinear problem. 
\begin{center} 
\includegraphics{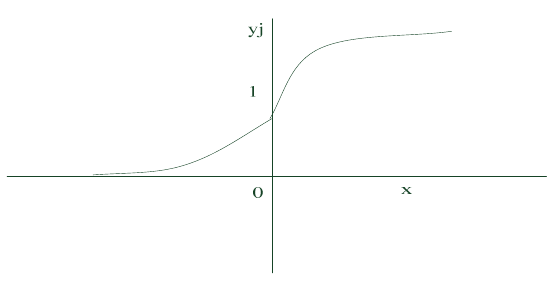}\\ 
Figure 5: A simple sketch of the sigmoid function. The function is differentiable, nonlinear and monotonically increasing. These properties make it an ideal activation function to use in ANN. 
\end{center} 
Mathematically, the function is expressed as: 
\begin{equation} 
f(x) = \frac{1}{1 + e^{-\sigma x}} 
\end{equation} 
The shape of the sigmoid function is being controlled by the constant value, $\sigma$ in the equation. The flexibility of the function is based on this constant value and function can take any value within the range of 0 and 1 or 1 and -1 (in hyperbolic tangent, $\tanh$). If large value is assign to $\sigma$ (between 0 and 1), the function will resembles the step function; and if small value of $\sigma$  is assign, the function approximate to a straight line.  
\subsubsection{Model's Learning Rate} 
Network learning, rate is a parameter that is use to control the speed at which ANN model learns. It is usually assign a value within the range of 0 and 1. When the learning rate is assigned a large value, the network learns very fast with high tendency of skipping the global minima (which is the ultimate desired solution). The global minimum is the lowest point where the error is minimal and the value of the weight vector at the point is returned as the optimum value. Similarly, if very small value is assign to the learning rate the learning process will be slowed and take long time before reaching the global minimum. 
\subsection{Learning Models in ANN} 
The first attempt to model the biological nerve cell was found in the work of McCulloch and Pitts ~\cite{icp21} in which a model based on the Threshold Logic Unit (TLU) was proposed. The TLU output signal (‘1’) when the summed weighted-input equals or exceeds certain threshold value, otherwise the model respond with ‘0’ as output. The Perceptron and the Multilayer Perceptron are the common models in ANN. 
\subsubsection{Perceptron} 
Earlier neural network models like the McCulloch and Pitt model are devoid of weight vectors (free parameter) that are responsible for the learning capability in the ANN model ~\cite{icp6}. The weight vectors can be tuned to learn relationship in the training data. The first model of Perceptron was proposed in 1958 by Rosenblatt ~\cite{icp13} to enhance McCulloch and Pitt model by including a dynamic parameter that can be adjusted to suit the requirements of the task data via training. In 1969, Minsky and Papert scrutinised the computational capabilities of Rosenblatt Perceptron and came up with a new approach toward solving linearly separable data ~\cite{icp11}. The work of these two scientists also highlighted the weaknesses of the Rosenblatt Perceptron of being able to only classify linearly separable data. Their findings affect the ANN research community in two major ways:\begin{itemize} 
\item firstly, it dispelled earlier believes that the Rosenblatt’s Perceptron is computationally universal; 
\item secondly, it resulted in inactivity of research in the field of artificial neural network for quite a long time ~\cite{icp21}\end{itemize}. Despite the limitation of the Perceptron to solving only linearly separable data, it is still regarded as one of the powerful artificial neural model ~\cite{icp17} and forms the building block for other powerful model (the Multilayer Perceptron). 
The basic elements of the Perceptron include: \begin{itemize} 
\item the \textit{input pattern} fed into the association unit where they are randomly paired with weight vectors 
\item \textit{an association unit} made up of positive and negative weight vector values ranging from 1 to -1.   
\item \textit{an output unit:} output from the association (product of inputs and weights) are summed in the summation unit (housed in the output unit). Output from the summation unit is used by the activation function (step function) which decide the net response (+1/-1 or 1/0). Based on this response the weights are modified and the process repeats. 
\end{itemize}  
 
Multilayer Perceptron and the Perceptron incorporate an extra node known as the \textit{bias node}. Bias node is an extra input node with fixed value that is added to the network. The inclusion of this bias node prevents having zero response when input vectors with zero values are encountered. This node is usually assign a constant value of -1 as input but the weight is assign using the normal procedure of weights assignments usually in randomised fashion for initialisation. 
\subsubsection{Perceptron Error Function} 
In ANN the errors generated during training a model form the basis of learning. These generated errors are used to update the network weight vectors which promote learning. Perceptron error function is given by: 
\begin{equation} 
E_p = \frac{1}{2} a (d - y)^2 
\end{equation} 
Where: 
\begin{itemize} 
\item $E:$ denotes the Perceptron error function; 
\item $d:$ denotes the desired output supplied with the training data in supervised learning technique; 
\item $a:$ is the activation function (such as the sigmoid function); and 
\item $y:$ is the network response and expressed as: 
\end{itemize} 
\begin{equation} 
y_j =  \Sigma^n _{i=0} w_{ij}x_i 
\end{equation} 
 
The object is to minimise the error $E_p $  and obtain an optimum value for weight vectors, $\textbf{w}$. Minimisation of the error is achieve through taken the partial derivative of the error function with respect to the weight vector (for each set of values of $w$, there are corresponding output and error. Errors decrease or increase depending on the weight values): 
\begin{align} 
\frac{\delta E}{\delta w_i} &= \frac{1}{2}e^2\\ 
\nonumber 
&=e\times\frac{\delta E}{\delta w_i}a(d -\Sigma w_ix_i)\\ 
\nonumber 
& = - e \times a^{'} (x) \dot{•}x_i 
\end{align} 
where  
$$a^{'}(x) = (d - \Sigma w_ix_i)$$ and $$y = \Sigma w_ix_i)]$$ 
 
\begin{align} 
& \Longrightarrow \frac{\delta E}{\delta w_i} = - e \times g^{'}(x).x_i  
\end{align} 
 
The network weight vectors are updated based on the error term; hence the weight update rule is given as: 
 
\begin{align} 
w_i \leftarrow w_i + \eta (d_j - y_j) . x_i 
\end{align} 
 
The constant is the learning rate that ensures a gradual modification of the weight vector [if set correctly]; with this term, only a fraction of the error function is being used in the weight update.  
 
\subsubsection{Gradient Descent (Delta Rule)} 
Gradient descent (also known as the \textit{delta rule}) is a supervisory approach aimed at minimising training error. Error minimisation is achieved through an iterative process of gradient descent algorithm that is based on the slope (the gradient of a function) of the error surface given as a function of the network synaptic weights. 
The Gradient Descent (GD) is the learning algorithm use in minimising error with the ultimate goal of reaching Global Minima (GM) in the error surface. This algorithm is based on a differential approach to error minimisation and guarantees convergence to global minima only if Local Minima (LM) are non-existent. However, this not always the case as the surface is likely to contain numerous regions of undesirable local minima affecting algorithm's performance ~\cite{icp17}. Local minima are regions in the error function where the gradient descent is unable to converge to global minima (see figure 6). These regions appear like valleys in the error surface hindering gradient descent to advance, hence resulting in high error that is sometime worse than the step function ~\cite{icp19}. The below figure depicts this phenomenon. 
\begin{center} 
\includegraphics{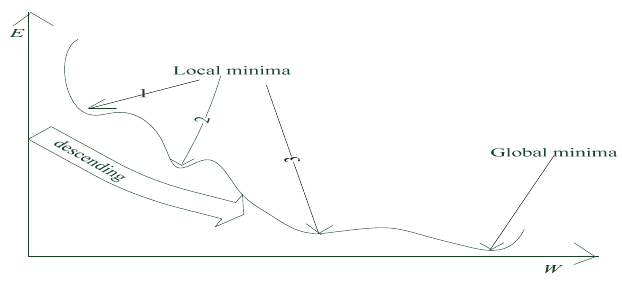}\\ 
Figure 6: A figure depicting local minima and global minima found in error function surface. The descending arrow only guarantees reaching the first valley (first local minimum). These regions of local minima will prevent converging to the global minima. 
\end{center} 
 
The ultimate goal of gradient descent algorithm is to reach the lowest region (global minima) in the error surface and path to this region is being obstructed by numerous local minimum. The local minima usually occur when the partial derivative of the error function is zero or it is too small but the error, e is still high ~\cite{icp24} i.e. 
\begin{align} 
\frac{\delta E}{\delta w} = 0 
\end{align} 
 
With a differentiable function, finding region with lowest error is made possible. The sigmoid function is the widely used function in this respect. A possible approach to avoiding the problem of local minima is to introduce constant momentum parameter that ensures the skipping of these regions of local minima ~\cite{icp12}. 
 
\subsubsection{The Feed-forward Multilayer Perceptron} 
The computational limitation of the Perceptron [since it is limited to only solving linearly separable data] prompted the development of MLP ~\cite{icp11}. The MLP is an artificial neural network model that is capable of solving both linear and nonlinear problems through the inclusion of additional layer on the network architecture. The learning mechanism of the MLP is known as the supervised backpropagation learning algorithm. Backpropagation algorithm functions in phases:\begin{itemize} 
\item forward phase; and  
\item backward propagation of error phase.\end{itemize} 
Figure 7 shows the typical architecture of the MLP network and how the phases involve in the backpropagation operation. 
\begin{center} 
\includegraphics[scale=0.5]{feedforward}\\ 
Figure 7: The architecture of a simple feedforward multilayer perceptron depicting the two execution phases. The forward phase computes the network response based on the weighted-input and the backward phase propagate the error generated at the output layer back into the network to update weight vectors accordingly. 
\end{center} 
While the MLP is able to solve both linear and nonlinear data mainly due to the inclusion of additional layers in the architecture known as the hidden layers; the inclusion of this layer substantially increase the network complexity and difficulty in deciding the weights responsible for the errors generated at the output layer. Since the weight vector(s) responsible for the error is not known, the error in the output layer is transmitted back into the network and the blame is shared amongst the network weights. The backpropagation learning algorithm is the formal description of the operations involve. Recently, there is significant improvement in multilayer models such as deep learning approach. Emerging approaches such as deep learning will be the focus of future article. 
\subsubsection{Multilayer Perceptron Error Function} 
Early neural network models [notably the TLU] are devoid of weight vectors responsible for the model learning. The Perceptron was successfully developed to incorporate weight vectors in the ANN model thereby improving on TLU performance ~\cite{icp16}, however, the weight vectors in the Perceptron model are independent of each other and there is no common factor that can influence their response. The absence of this common or unifying factor is responsible for the inability of the Perceptron to classify nonlinear data. The architecture of the MLP and the supported dependency among weight vectors in the hidden layers has led to having a common way of influencing their collective response. This is so because the weight vectors in the MLP architecture are related, hence their responses can be influenced. Figure 7a presents a simple segment of MLP that shows how the hidden layer weight vectors and the input layer weight vectors share something in common. 
\begin{center} 
\includegraphics[scale=0.5]{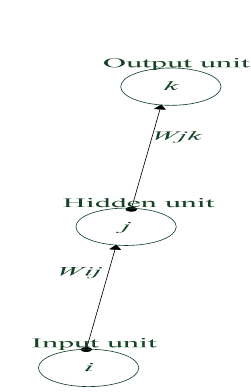}\\ 
Figure 7a: Segment of MLP to illustrate single connection from input to hidden then to output units. 
\end{center} 
 
The MLP error function is use in computing the difference between the desired output, $d$ and the actual output, $y$. MLP uses the sigmoid function to transform input to output between $-\infty$ to $\infty$  in the range of 0 and 1. Sigmoid function guarantees that the output is bounded within 0 and 1. The error function is given by: 
\begin{align} 
E_m = \frac{1}{2} a ( d - y)^2\\ 
\nonumber 
y_j = \Sigma^n _{i=0} w_{i,j}x_i\\ 
\nonumber 
e =  a (d - y)^2 
\end{align} 
The manner in which the weight vectors are updated is similar to that of the Perceptron and the update starts from the output layer back into the network using the following equations: 
\begin{align} 
w_{j,k} \leftarrow w_{j,k} + \eta a_j 
\end{align} 
where the hidden layer activation function given by: 
\begin{align} 
\nonumber 
a_j = f(in_j) 
\end{align} 
and output unit activation given by $ a_k =  f(in_k)$ and $ in_k = \Sigma _{j} w_{j,k}a_j$, hence the update rule will now be : 
\begin{align} 
w_{j,k} \leftarrow w_{j,k} + \eta a_j \times \Delta_k 
\end{align} 
  
where $\Delta _k = e_k f^{'} (in_k)$ and the error at the hidden unit, given by j: $ \Delta_j = f^{'}(in_k) \Sigma_{k}w_{j,k}\Delta_k $ 
This is the point at which the backpropagation algorithm is needed, the term $ \Sigma_{k}w_{j,k}\Delta_k $ is the error function at the output unit, $k$, this error is propagated back into the network, each hidden unit receive its proportion based on its weight contribution toward the generation of error in the output unit, $k$. 
update rule at the hidden unit is given by: 
\begin{align} 
w_{i,j}\leftarrow w_{i,j} + \eta a_i \times \Delta_j 
\end{align}   
The algorithm of back-propagation will be presented later. 
\subsubsection*{Data: Linear and Non-Linear} 
\textit{Linear separable data} is any data that is believed to be linearly separable and can be solved using well defined relationship such as the of straight line equation in 2-dimensional space otherwise it is non-linear. Straight line equation given by:  
\begin{equation} 
y = mx + c 
\end{equation} 
Where $y$ denote the output (dependent variable); $m$ denote the gradient; and $c$ denote the $y-intercept$ and $x$ the input as the independent variable. 
The operations of the TLU [and the Perceptron] are based on the concept of straight line equation. Consider a two-dimensional input data and weight vectors; the summation of weighted-inputs that equals zero separates the input data into two distinct classes. The next equation resembles the previous straight line: 
\begin{equation} 
w_1 x_1 + w_2 x_2 = \theta 
\end{equation} 
     [$\theta$ denotes the threshold] 
In this equation, the weighted-inputs are linearly combined which made it possible to be separated with the aid of straight line equation concept. Input patterns of this form can be solved using simple TLU or the Perceptron. 
\begin{center} 
\includegraphics[scale=0.5]{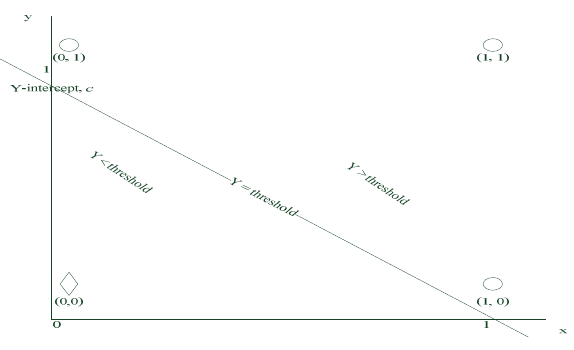}\\ 
Figure 8: The Illustration of linear separation in 2D plane. The point at which $y = threshold$ is the decision line that determine the boundary between the two distinct classes. The point at which the activation equals the threshold is the critical condition for classification; one side of the space $y > \theta$ (threshold) represents class ‘1’ and the other side $y <\theta$ represent class ‘0’. 
\end{center} 
 
\section{Learning Algorithms} 
\subsection{Back-propagation algorithm} 
The backpropagtion learning algorithm is an application of a statistical method known as stochastic approximation originally proposed in 1951 ~\cite{icp12}. The stochastic nature of the algorithm often leads to slow convergence. The algorithm is also based on the gradient descent search algorithm.\\ 
\textbf{Algorithm 1: The Backpropagation Learning Algorithm ~\cite{icp21}}\\ 
\textit{Steps:} 
\begin{enumerate} 
\item \textbf{Initialisation}: $w \leftarrow randomvector()$ 
\item \textbf{Training}: for training example $(x, d)$, repeat: 
\begin{enumerate} 
\item $h_j =  \Sigma _{i} x_iv_{i,j} $ 
\item $a_j: g(h_j) =\frac{1}{1+e^{- \beta h_j}}$ 
\item until: $h_j =  \Sigma _j a_j w_{j,k}$ 
\item $y_k: g(h_k) = \frac{1}{1+e^{- \beta h_k}}$ 
\item $\delta_{o,k} = (t_k -y_k)y_k (1 - y_k)$ 
\item update hidden layer: $\delta_{h,j} = a_j (1 - a_j)\Sigma_k w_{j,k} \delta _{o,k}$ 
\item update weigths: $w_{i,j}\leftarrow w_{i,k} + \eta \delta_{o,k}a_j$ 
\item $w_{i,j} \leftarrow w_{i,j} + \eta (d_j - y_j).x_i$ 
\end{enumerate} 
\item untile learning stop 
\end{enumerate} 
\textbf{Visualising the operation of the basic steps in back-propagation algorithm} 
\begin{center} 
\includegraphics[scale=0.5]{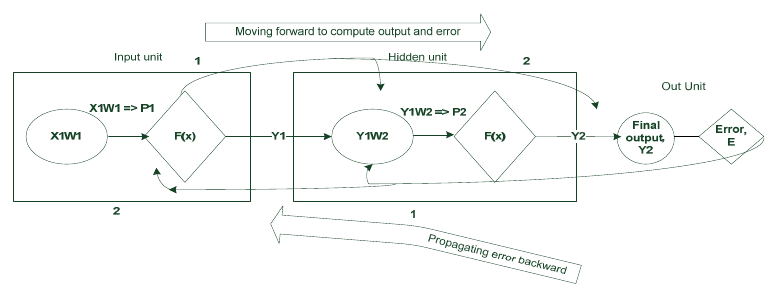}\\ 
Figure 11: Visual model of the backpropagation algorithm operation. 
\end{center} 
\textit{Some notations:} 
\begin{itemize} 
\item $p_i$: product of weighted inputs 
\item $d$: desired output 
\item $y_i$: actual output 
\item $E$: network error 
\item $x_i$: network input 
\item $w_i$: weight vectors. 
\end{itemize} 
The final output,$y_2$ from the model depends on the network free parameter $w_i$ and the partial derivative of the error with respect to $w_i$ given by: 
\begin{align} 
\frac{\delta E}{\delta w_i} = \frac{\delta E}{\delta y_2}\dot{•} \frac{\delta y_2}{\delta w_i} 
\end{align} 
The forward phase computes the net final output and expressing the network error in terms of the immediate variables between the network error and the first weight vector in the model [$w_2$ in this respect] to share the network error [blame] among the weight vectors.\\ 
\textit{Contribution of $w_2$ toward the network error, $e$:} 
\begin{align} 
\frac{\delta E}{\delta w_2} = \frac{\delta E}{\delta y_2}\dot{•} \frac{\delta y_2}{\delta w_2} 
\end{align} 
 
\begin{align} 
\frac{\delta E}{\delta y_2} = - (d - y) 
\end{align} 
with: 
\begin{align} 
e = \frac{1}{2}(d - y)^2 
\end{align} 
 
\textit{Applying chain rule to expand the expression} $\frac{\delta y_2}{\delta w_2}$ 
 
\begin{align} 
\frac{\delta E}{\delta w_2} = \frac{\delta E}{\delta p_2}\dot{•} \frac{\delta p_2}{\delta w_2} 
\end{align} 
 
$$\frac{\delta p_2}{\delta w_2} = y_1$$ and 
 $$\frac{\delta y_2}{\delta p_2} = y_2(1-y_2)$$ 
  
 \textit{Contribution of $w_1$ toward the network error, $e$:} 
\begin{align} 
\frac{\delta E}{\delta w_1} = \frac{\delta E}{\delta y_2}\dot{•} \frac{\delta y_2}{\delta w_1} 
\end{align} 
 
\textit{Applying chain rule to expand the expression} $\frac{\delta y_2}{\delta w_1}$ 
\begin{align} 
\frac{\delta y_2}{\delta w_1} = \frac{\delta y_2}{\delta y_1}\dot{•} \frac{\delta y_1}{\delta w_1} 
\end{align} 
The process continuous and the chain rule is continuously applied in evaluating the expression. 
 
\subsection{Perceptron} 
\begin{center} 
\includegraphics[scale=0.5]{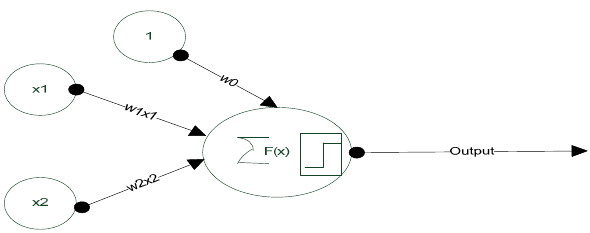}\\ 
Figure 12: Perceptron network with set of weighted-inputs, processing unit and the output unit. The linear sum of the inputs $x_1, x_2$ and the bias node and their corresponding weights is denoted by the summation of $w_{i,j}x_i$. 
\end{center} 
The perceptron error function is based on the differences between network actual output and the expected output. The goal is to minimise the error and obtain an optimum value for weight vectors, $w$; the minimisation of the error is achieve through taken the partial derivative of the error function with respect to the weight vector. The error function of perceptron given earlier as: 
\begin{align} 
\nonumber 
E_p = \frac{1}{2}a(d - y)^2 
\end{align} 
The network weight vectors are updated based on this error term and the Perceptron algorithm. The weight update rule is given as: 
\begin{align} 
w_i \leftarrow w_i + \eta (d_j - y_j) \dot{•} x_i 
\end{align} 
If set correctly, the constant, $\eta$ [the network learning rate] ensures a gradual modification of the weight vector; with this term, only a fraction of the error function is being used in the weight update rule. 
 
\subsection*{Sensitivity Analysis for Training Parameters} 
Sensitivity analysis refers to the examination of how different values of an independent variable impact other dependent variable under the same set of conditions. The reason for carrying out this analysis is because the neural network relies on free parameters [parameters that have to be set by the user in order to facilitate the ability of the ANN to correctly recognise] in order to produce good result with the minimal error, the use of sensitivity analysis to determine optimum values for these parameters will ensure the attainment of good performance. Failure to determine the right settings for these parameters will affect the network's performance in the following ways: 
\begin{itemize} 
\item the network will take longer time to learn during training [if at all it will learn]; or 
\item causes the oscillation of the weights vectors during the network training. 
\end{itemize} 
This technique is often employed in building effective and robust model. Models are evaluated on range of parameters values to ascertain the best set to be used by the model in subsequent tasks. Well-tuned parameters for optimum performance (parameters setting leading to lowest error in the shortest time). The main reason for the analysis is for network optimisation and often achieve by employing techniques such as grid search or randomised search. Detail of these techniques in future article.

\section{Brief on Vectors and Modelling} 
Vectors are extensively used in ANN and play vital role in describing input patterns and simplify the difficulty in visualising concepts. Unlike scalar quantities which have only magnitude, vector quantities have both magnitude and direction. Vectors can be used to perform simple mathematical operations such as addition, subtraction and multiplication to more complex operations involving numerous variables. Vectors are represented in the Cartesian coordinate system [a rectangular system that identifies the coordinates of vectors]. This system of Cartesian coordinate describes the vectors through the representation of their respective coordinates. In relation to ANN, vectors are quite useful in describing the behaviour of neuron in pattern space ~\cite{icp16}. Pair of number (components of vector) is used to describe a vector in a two-dimensional coordinate system. Common example in ANN is the representation of weight vector in the form of: 
\begin{align} 
W = w_1, w_2,w_3, w_4, w_5, .... ,w_n 
\end{align} 
and the input vecor space: 
\begin{align} 
X = x_1,x_2,x_3,x_4,x_5, .... x_n 
\end{align} 
 
\subsection{Vectors comparison} 
An angle separates two vectors in a $2-dimensional space$. The combination of separating angle between two vectors and their corresponding lengths defines their ‘inner product’ as shown below: 
\begin{align} 
W.X = ||w||\dot{•}||x||\cos \theta 
\end{align} 
where:  $||w||$ denote the length of the weight vector and $||x||$ denotes the length of the input vector. 
The generalisation of $W.X$ in $n-dimensions$ is given by: 
\begin{align} 
w.x = \Sigma^n _{i=1} w_ix_i 
\end{align} 
The dot product $w.x$ of vectors is always a number and this number determine the direction at which the vector points, a geometrical illustration is given in figure 9: 
\begin{center} 
\includegraphics[scale=0.5]{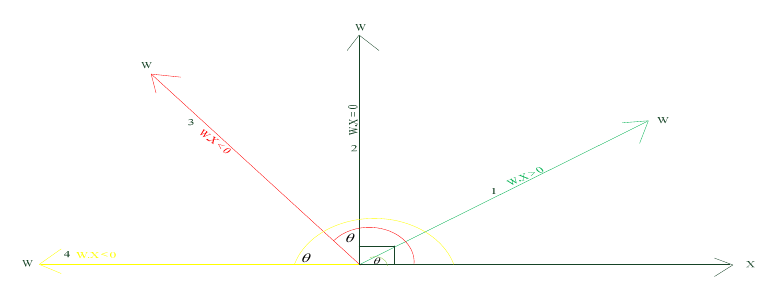}\\ 
Figure 9: Comparing vectors through their inner products. The vector X remains fixed while the direction of vector $W$ is varied across the system. 
\end{center} 
 
It can be discerned from the figure that the vectors points in the same direction when the dot product is positive. Similarly, the vectors points in the opposite direction if the dot product is negative. A special case is when the dot product equals zero; at this point the vectors are orthogonal. In all these scenarios, the vectors’ direction depends on the angle, . The closer the vectors, the higher their dot product value and the farther they are the more negative the dot product. The dot product of the two vectors in 2D resembles the TLU activation functions. The next figures illustrates vectors projection: 
\begin{center} 
\includegraphics[scale=0.5]{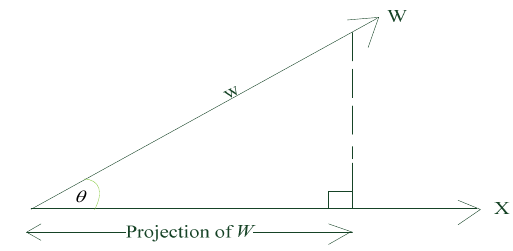}\\ 
Figure 10: The projection of vector $w$ onto the vector $x$ projection. 
\end{center} 
The projection of vector $w$ onto the vector $x$ is define as: 
\begin{align} 
\cos \theta = \frac{w_x}{||w||}
\end{align}

Multiplying the right hand side of the previous equation with $\frac{||x||}{||x||}$  [which is essentially equivalent to multiplying by the constant integer ‘1’] 
\begin{align} 
w_x = \frac{||w||.||x||}{||x||}\cos \theta 
\end{align} 
with $||w||.||x||\cos \theta = W.X$ 
\begin{align} 
{\Longrightarrow} {w_x} = \frac{w.x}{||x||} 
\end{align} 
Vectors are useful in establishing connection between the function of threshold logic unit (TLU) and linearly separable data and independent of the dimensionality of the pattern space ~\cite{icp16}. 
 
\subsection{Modeling} 
Modeling is an act of simplifying any situation through the identification of relevant variables in physical world. Through modeling, the behaviour of system can be well understood and accurately predicted. Modeling comes in different varieties depending on the situation and area of application ~\cite{icp27}. Models could be tangible [models that could be seen and touch] or intangible/conceptual models ~\cite{icp26} that only exist in the human mind which are understood only when translated into visual models. 
 
\subsubsection{Conceptual Modeling} 
Conceptual modeling is a form of modeling that only exists in the mind individual and useful in understanding the situation at hand. In the concept of software design process, conceptual modeling involves the organisation of the system to be developed in terms of the components that make up the system and their relationships ~\cite{icp25}. Unified Modeling Language (UML) is a form of graphical modeling language that provides the structural and behavioral description of software design process. UML uses notations to illustrate concepts and present system’s architecture in the form of context diagram, class diagram, sequence diagram, flow-chart etc. 
 
\subsubsection{Mathematical Modeling} 
Mathematical modeling is used in providing a description of the system that will be developed using mathematical terminologies. Mathematical modeling can take many forms and the form of mathematical modeling considered in ANN comes in the form algorithms and functions, etc.

\section{Conclusion}
This section concludes the first part of the article. Part 2 of the article will specifically focus on relevant theoretical concepts in the widely use machine learning and deep learning models. 

\cleardoublepage

\addcontentsline{toc}{section}{References}
\bibliographystyle{plain}
\bibliography{intellcomp}

\end{document}